\documentclass[11pt]{article}
\usepackage{nodalida2025}
\usepackage{times}
\usepackage{url}
\usepackage{latexsym}
\usepackage{graphicx}
\usepackage{hyperref}
\usepackage{caption}

\aclfinalcopy 

\title{Towards Multilingual LLM Evaluation for Baltic and Nordic languages: A study on Lithuanian History}

\author{Yevhen Kostiuk$^{1,2}$\\
$^1$ARG-tech, \\
University of Dundee \\
$^2$OpenBabylon \\
{\tt ykostiuk001@dundee.ac.uk} \\\And
  Oxana Vitman \\
  University of Bremen  \\\And
  \L ukasz Gaga\l a \\
  Georg-August \\ Universität Göttingen \\\And
  Artur Kiulian \\
  OpenBabylon}
 
\date{}

\begin{document}
\maketitle
\begin{abstract}

In this work, we evaluated Lithuanian and general history knowledge of multilingual Large Language Models (LLMs) on a multiple-choice question-answering task. The models were tested on a dataset of Lithuanian national and general history questions translated into Baltic, Nordic, and other languages (English, Ukrainian, Arabic) to assess the knowledge sharing from culturally and historically connected groups. We evaluated GPT-4o, LLaMa3.1 8b and 70b, QWEN2.5 7b and 72b, Mistral Nemo 12b, LLaMa3 8b, Mistral 7b, LLaMa3.2 3b, and Nordic fine-tuned models (GPT-SW3 and LLaMa3 8b).  

Our results show that GPT-4o consistently outperformed all other models across language groups, with slightly better results for Baltic and Nordic languages. Larger open-source models like QWEN2.5 72b and LLaMa3.1 70b performed well but showed weaker alignment with Baltic languages. Smaller models (Mistral Nemo 12b, LLaMa3.2 3b, QWEN 7B, LLaMa3.1 8B, and LLaMa3 8b) demonstrated gaps with LT-related alignment with Baltic languages while performing better on Nordic and other languages. The Nordic fine-tuned models did not surpass multilingual models, indicating that shared cultural or historical context alone does not guarantee better performance.


\end{abstract}

\section{Introduction}
Large Language Models (LLMs) provide a functional framework for tackling various natural language processing (NLP) tasks, such as question-answering~\citep{izacard_few-shot_2022, dong2024understand}, machine translation~\citep{Zhu2023MultilingualMT, kocmi-etal-2024-findings} and so on. 
However, LLMs have shown less reliable results for low-resource languages~\citep{ranjan2024comprehensivesurveybiasllms, sakib2024unveilingmitigatingbiaslarge} due to the smaller fraction of available data in comparison to English and a few other widely spoken languages.

Benchmarking multilingual LLMs across languages is essential for evaluating their capabilities. However, the availability of high-quality, culturally aligned datasets remains a challenge. This need for culturally aligned high-quality datasets becomes even more critical when evaluating historical knowledge, where ensuring linguistic and cultural fairness adds a layer of complexity.

Verifying comparability of the results on historical knowledge QA datasets requires that a single set of historical events is queried in all languages. The choice of that set is likely to be biased to events more represented in widely spoken languages. Conversely, events that are more region- or cultural-specific are less likely to occur in the benchmarks. Finding and addressing these gaps is important to improve the fairness of LLMs and highlight historical and cultural biases.

In this work, we focus on evaluating multilingual LLMs on Lithuanian and general history. Our goal is to determine how LLMs perform on Lithuanian history exam questions when prompted in different languages and explore the alignment between languages and historical awareness, particularly within the Nordic and Baltic language groups.

Our contribution is the following:
\begin{itemize}
    \item We automatically translated publicly available Lithuanian history exam question-answering dataset into Nordic (Danish, Finnish, and Swedish), Baltic (Estonian and Latvian), and other (Arabic, Ukrainian, and English) languages and partially manually evaluated it.
    \item We tested GPT-4o~\citep{openai2024gpt4technicalreport}, LLaMa3.2 3b, LLaMa3 8b, LLaMa3.1 8b and 70b~\citep{dubey2024llama3herdmodels}, Mistral Nemo 12b~\citep{jiang2023mistral7b}, QWEN2.5 7b and 72b~\cite{qwen2.5, qwen2}, and GPT-SW3 and Nordic-trained LLaMa3 8b~\citep{ekgren2023gpt} models and compared their achieved accuracy scores per language and its average per language group.
\end{itemize}

Our findings revealed that GPT-4o consistently outperformed other models across all evaluated languages and language groups on a dataset of LT-related and general history questions. Larger open-source models, such as LLaMa3.1 70b and QWEN2.5 72b, also demonstrated strong and consistent performance in all languages. In contrast, smaller models like Mistral Nemo 12b, LLaMa3 8b, LLaMa3.2 3b, and LLaMa3.1 8b showed notable gaps in their historical knowledge from a Lithuanian perspective, particularly with Baltic languages, despite Lithuanian being part of this group. The best performance was observed in the Nordic language group, suggesting that cultural or historical alignment alone does not ensure higher accuracy. Interestingly, the Nordic pre-trained models failed to surpass the multilingual model.


\section{Related Work}

Pre-trained LLMs have exhibited a remarkable ability to encode and retrieve factual and common knowledge across different languages~\citep{wang2023cross, zhao2024tracing}. However, there is a notable variation in model performance across languages, with a strong shift toward high-resource languages~\citep{qi2023cross}, particularly languages with Latin scripts~\citep{ifergan2024beneath}. 

The datasets used for benchmarking multilingual LLMs are created using either one of the two approaches: human annotation~\citep{kocmi-etal-2023-findings,  goyal-etal-2022-flores} or translating existing annotated datasets using LLMs~\citep{lai-etal-2023-okapi}. 

Although datasets created by human annotators provide accurate translations and task-specific precision, they require considerable investment of both time and finances~\citep{yang-etal-2019-paws}.

On the other hand, with an advancement of LLMs, the translation performance of automatic tools has been significantly boosted lately. For example, ChatGPT demonstrates fewer errors with the launch of the GPT-4 engine, even for distant languages~\citep{jiao2023chatgpt}. The quality control research on the DeepL translation tool found that DeepL\footnote{https://www.deepl.com/} performed well in terms of translation accuracy, fluency, and naturalness, reaching an overall semantic similarity score 94.13~\citep{linlin2024artificial}.

This improvement elevated the creation of benchmark datasets on various tasks. DeepL was used for creating the X-FACT multilingual factual knowledge dataset translated in 25 languages~\citep{gupta-srikumar-2021-x}. In the research~\citep{thellmann2024towards}, five well-known datasets of various tasks were translated by DeepL into 21 European languages. LLMs with different numbers of parameters were evaluated on newly introduced datasets. The authors observed that models generally achieve higher performance on Romance and Germanic languages compared to Slavic languages.

ChatGPT was utilized to translate the 158K English instructions into 26 languages, including 7 low-resource languages~\citep{lai-etal-2023-okapi}. The data was used to instruction-tune LLM for multiple languages using reinforcement learning from human feedback. The resulting framework, Okapi, was also evaluated on datasets translated by ChatGPT from English into 26 selected languages. 

Relying on the aforementioned approaches, in our work we utilized ChatGPT and DeepL to create a dataset with special consideration on Baltic and Nordic languages. To ensure the quality of translation for each language, 100 samples of English version and it's corresponding translation were manually evaluated by native speakers. On average, 77\% of such language pairs were approved.

We offer evaluation on Lithuanian history exam tasks, assessing the ability of multilingual LLMs to exhibit particular cultural and historical knowledge and evaluating the alignment of this knowledge within regional language groups.

\section{Methodology}

In this paper, we investigate performance consistency of LLMs within Nordic and Baltic language groups on the Lithuanian history exams questions. We hypothesized that the LLMs perform better in this domain, when presented with questions in languages from Nordic and Baltic groups than from other due to the cultural, linguistic and historical similarities.

\begin{figure}
    \centering
    \includegraphics[width=\linewidth]{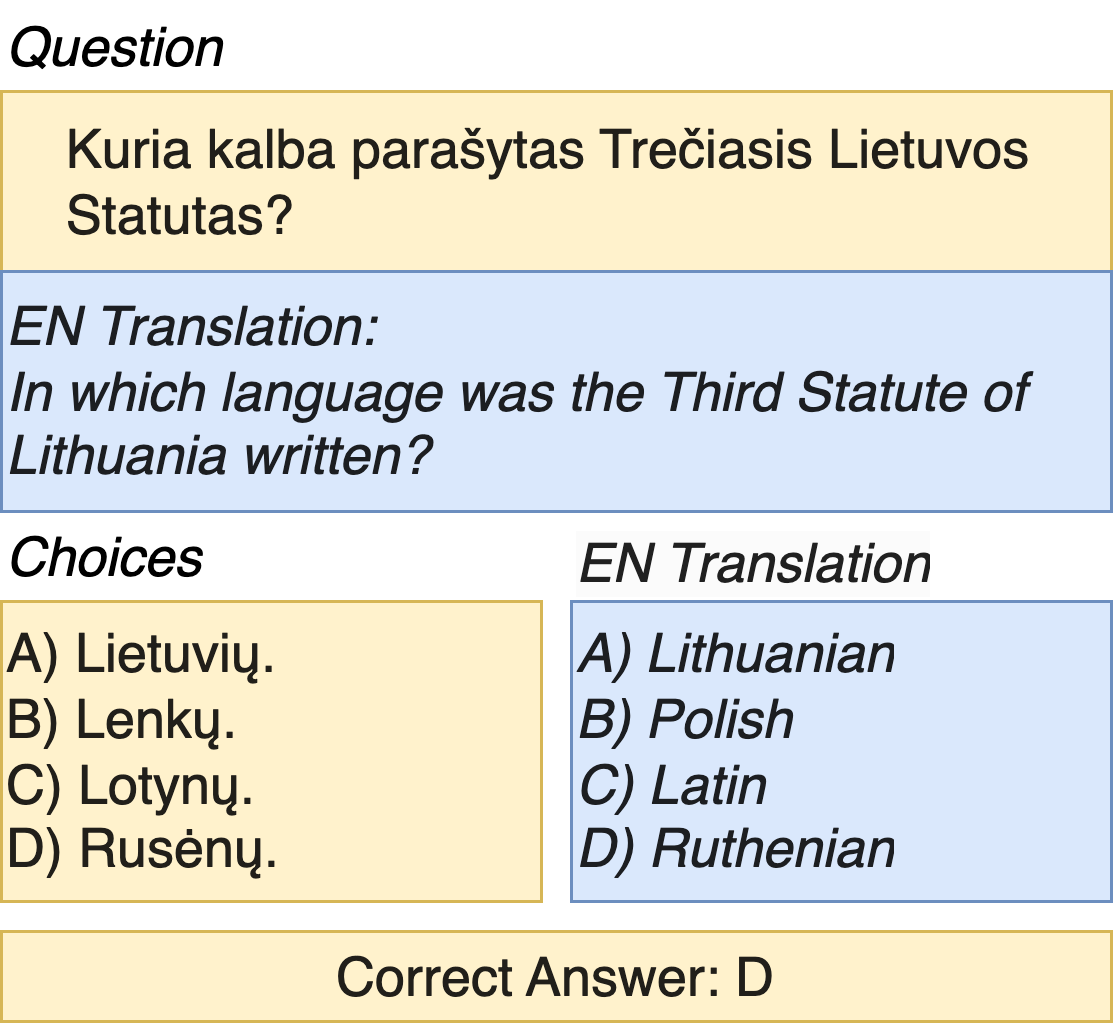}
    \caption{Example of the dataset sample in Lithuanian.}
    \label{fig:example_ds}
\end{figure}


The methodology consists of two steps: \textit{data preparation} and \textit{models' benchmarking}.

\paragraph{Data Preparation.} To test the hypothesis, we chose EXAMS~\citep{exams-dataset} dataset. Specifically, we used samples that correspond to Lithuanian history. Each sample contains a question, four different answer choices marked with the labels A,B,C and D with an indication of the correct one (see Figure~\ref{fig:example_ds}). Questions and choices are in Lithuanian. We manually removed the questions that require an image to answer it, obtaining 550 samples.

The dataset was machine translated into Nordic (Danish, Finnish, and Swedish), Baltic (Estonian and Latvian), and outside of Nordic-Baltic, multilingual language group: Ukrainian, English, and Arabic. In more details, the dataset was translated from Lithuanian to English, and then the English translations were translated in other languages. We used GPT-4o~\citep{openai2024gpt4technicalreport} and DeepL as translation algorithms, as they are proven to have a good machine translation performance from- and to-English rather than between underrepresented languages~\citep{wang2024exploring, hendy2023good}.

After that, we separated dataset into 2 parts: Lithuanian national history related questions (LT-related) and general history questions. We assigned a question to the LT-related group if it specifically mentions Lithuania, mentions Lithuanian historic figure or a question about the country that Lithuania was a part of or occupied by (e.g. Polish-Lithuanian Commonwealth, USSR after 1940 etc.). Other questions were assigned to a general history questions group.

To ensure quality, a subset of the dataset was evaluated manually by a group of native speakers. Annotators were presented with 100 English and translated language pairs (50 from LT-related and 50 from General history dataset) with 20 samples being the same for all the annotators to measure the annotators' agreement. For more details, see Appendix~\ref{sec:lt-en-tr}.

\paragraph{Models' benchmarking.} We experimented on the following models: LLaMa3 8b, LLaMa3.1 70b and 8b~\citep{dubey2024llama3herdmodels}, LLaMa3.2 3b~\citep{dubey2024llama3herdmodels}, Mistral Nemo 12b~\citep{jiang2023mistral7b}, GPT-4o~\citep{openai2024gpt4technicalreport}, QWEN2.5 7b and 72b~\cite{qwen2.5, qwen2}, and families of instruct pre-trained models developed by AI Sweden\footnote{\url{https://huggingface.co/AI-Sweden-Models}}: GPT-SW3~\cite{ekgren2023gpt} (126m, 356m, 1.3b, 6.7b) pre-trained for and LLaMa3 8b fine-tuned for Swedish, Norwegian and Danish. GPT-SW3 models were pre-trained for Swedish, Norwegian, Danish, Icelandic, English, and programming code and LLaMa3 8b (we will refer to this model as NRD LLaMa3 from now on to avoid confusion) was fine-tuned for Swedish, Norwegian and Danish. 

\begin{figure}
    \centering
    \includegraphics[width=\linewidth]{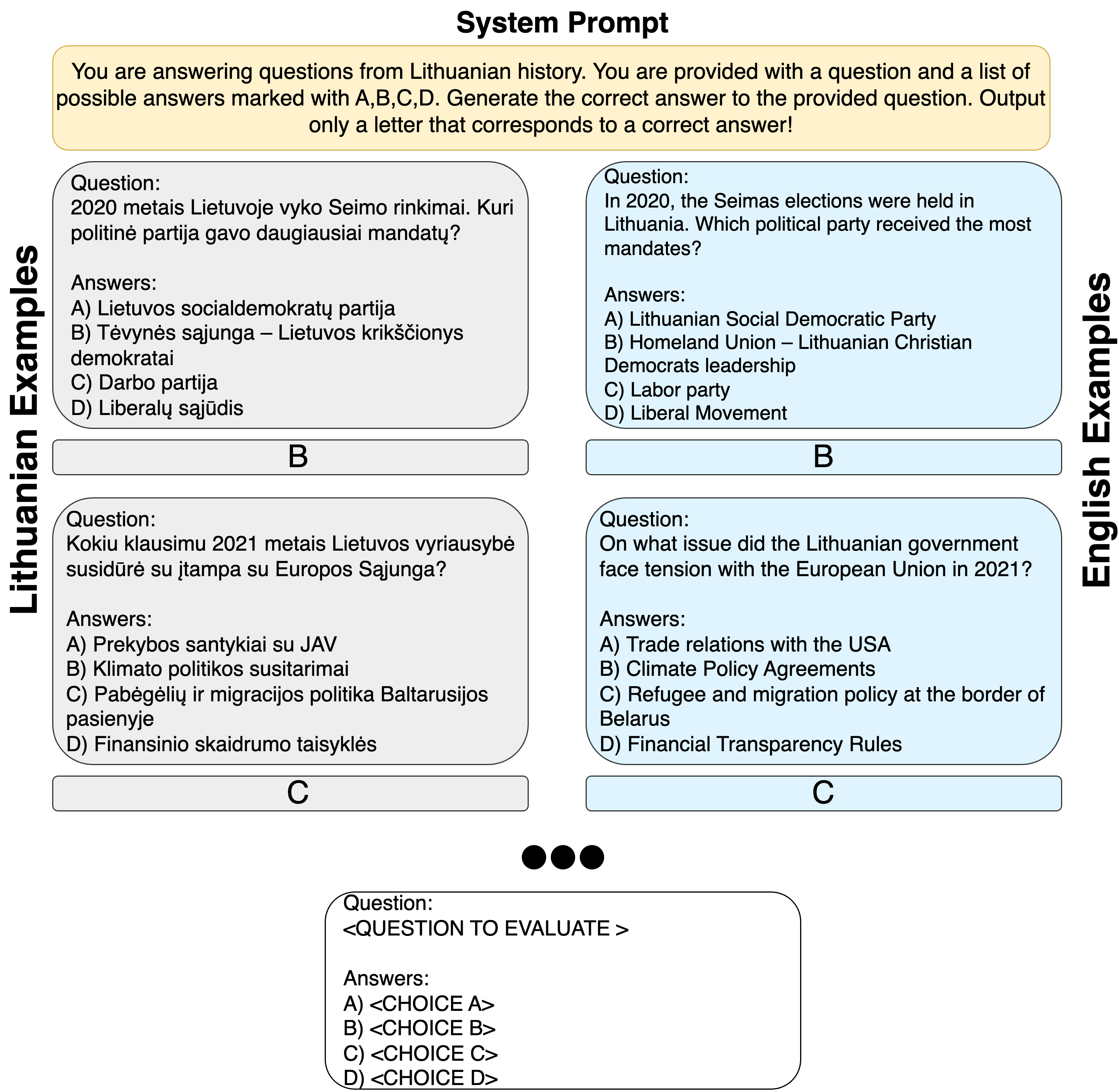}
    \caption{Example of the chat prompts that the model was presented to evaluate the dataset in Lithuanian and English languages.$ <\dots>$ is the actual question that the model is evaluated on.}
    \label{fig:prompts}
\end{figure}

In our experiments, we used multilingual instruct LLMs. During generation, all the parameters were set to defaults, except for random seed, which we set to 2 with Ollama~\footnote{\url{https://github.com/ollama/ollama}} framework for open source models. For Nordic models, we used implementation from the \textit{transformers}~\cite{wolf-etal-2020-transformers}. Specifically, we used GPT-SW3 and Swedish LLaMa3 8b\footnote{\url{https://huggingface.co/collections/AI-Sweden-Models/}}.

For each language in the translated dataset, the model was evaluated on the multiple choices question-answering task. The model was presented with a system message in English explaining the task, four question-answering examples in the corresponding language, and finally, a question with answer choices that the model has to answer. The examples were presented in the same format as the final question and consisted of question, four answer choices marked with A, B, C, D, and the correct letter for an answer as an expected output. Examples were taken from the modern (later than 2020) history of Lithuania, and do not intersect with questions in the dataset. Everything, except the system prompt, was presented to the model in the evaluated language (see Figure~\ref{fig:prompts}).

The results were parsed in the following way. If the model generated more than one letter, the generated text was separated into words. From these words, only capital letters were kept that corresponds to possible choice letters A,B,C, or D. If only one letter was present, it was considered as a final output. Otherwise, we assume that the model failed to produce a reasonable output and record it as if the model's answer was incorrect. As a result, we measured accuracy score for each model and each language.

\section{Results and Discussion}
\begin{figure}[]
    \centering
    \includegraphics[width=\linewidth]{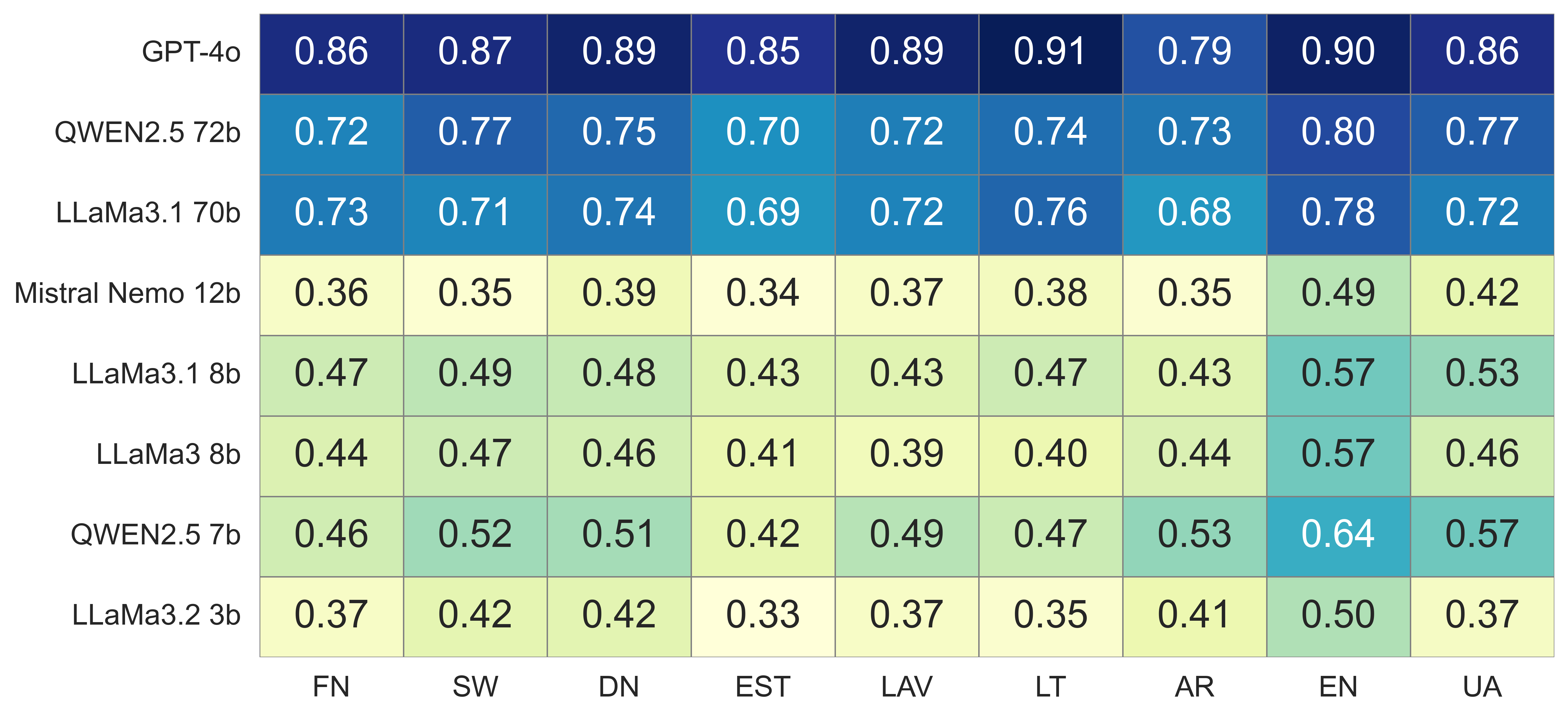}
    \caption{Accuracy results per language for LT-related history questions.}
    \label{fig:lt_res}
\end{figure}

\begin{figure}[]
    \centering
    \includegraphics[width=\linewidth]{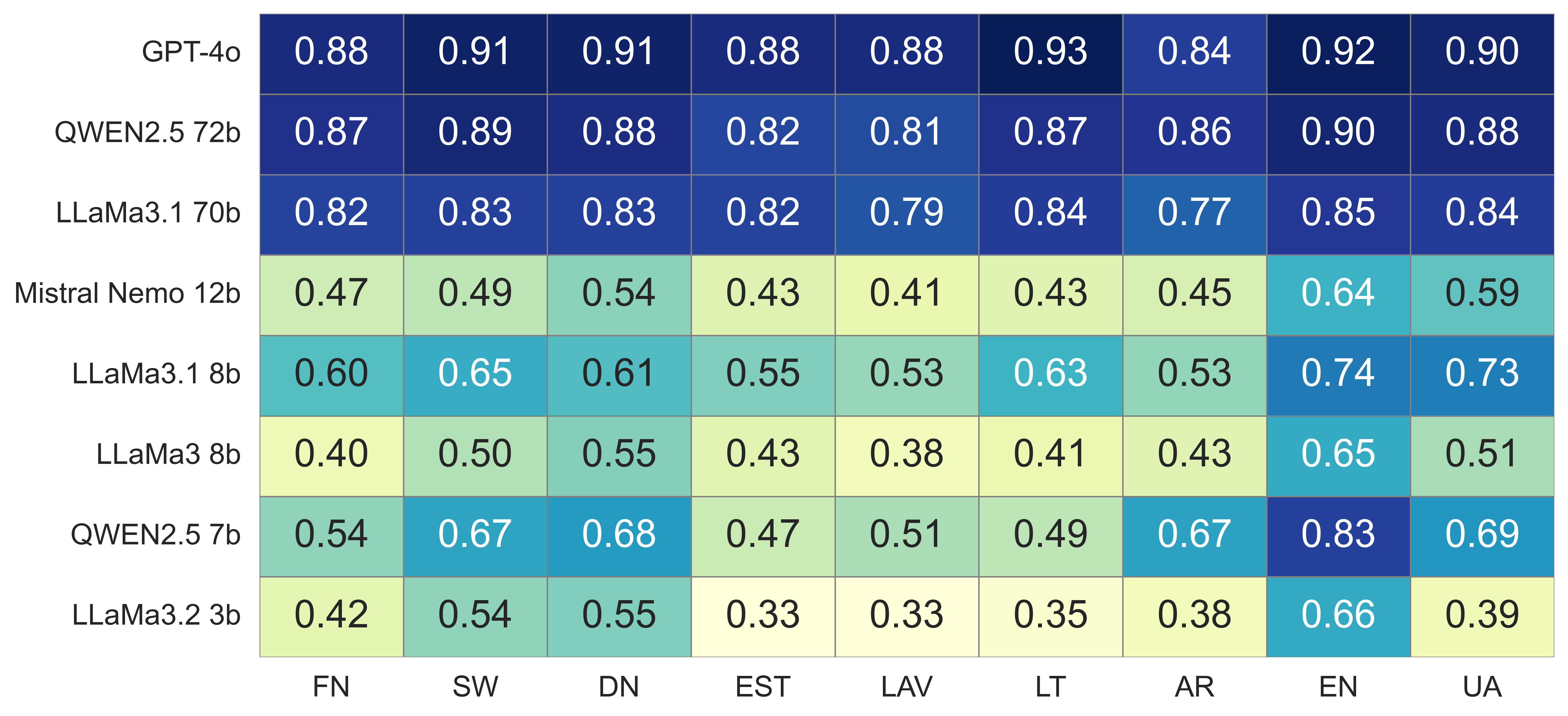}
    \caption{Accuracy results per language for general history questions.}
    \label{fig:gen_res}
\end{figure}

\begin{figure}[]
    \centering
    \includegraphics[width=\linewidth]{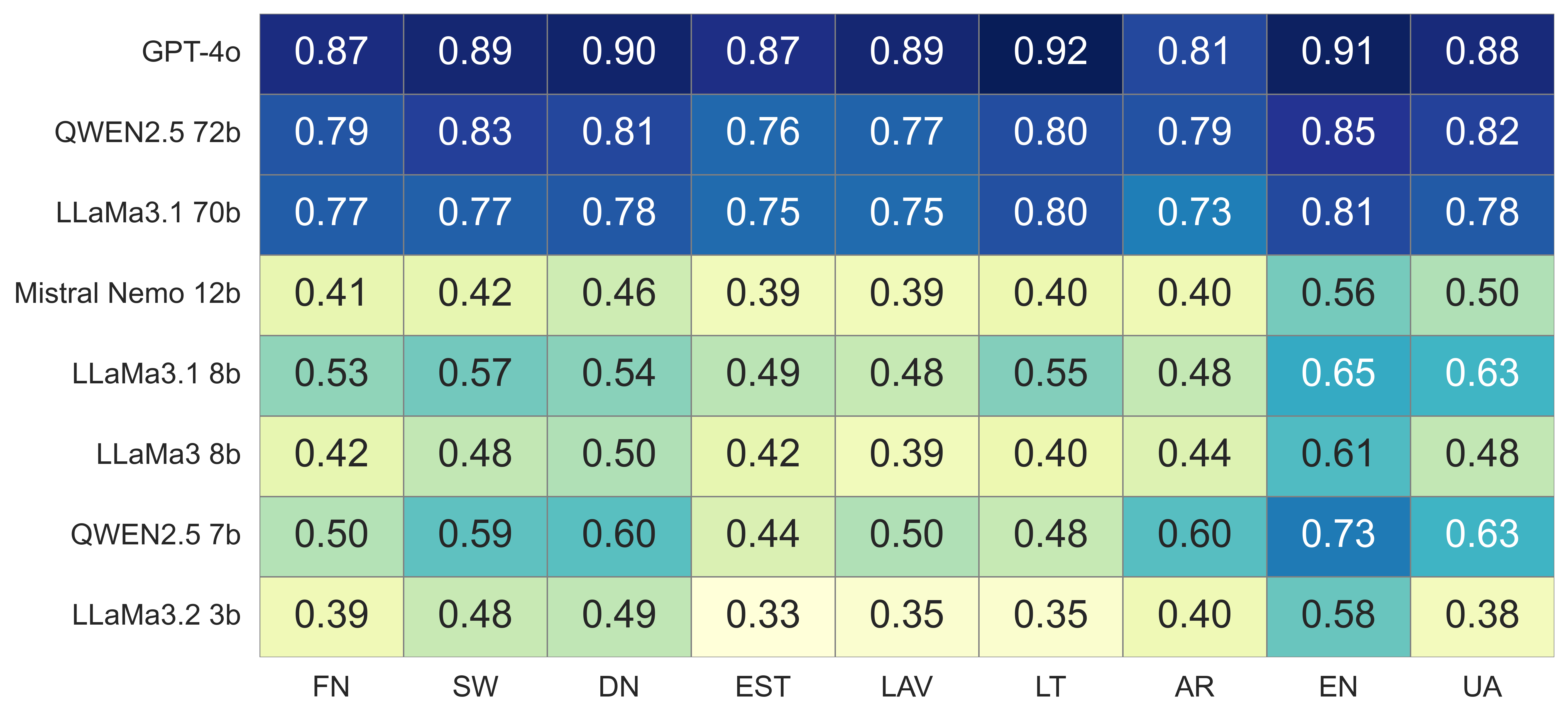}
    \caption{Accuracy results per language for merged LT-related and general history questions.}
    \label{fig:mer_res}
\end{figure}

For each LLM, we grouped its results per language group into Nordic (Danish, Finnish, and Swedish), Baltic (Lithuanian, Latvian, Estonian), and multilingual (Ukrainian, English, and Arabic). The accuracy scores per language and averages scores per language group are presented in Tables~\ref{tab:avg_results} and~\ref{tab:nordic_models_res} and on Figures~\ref{fig:lt_res}, \ref{fig:gen_res}, and~\ref{fig:mer_res}.

\begin{table*}[!h]
\centering
\begin{tabular}{l|ccc|ccc|ccc|}
\cline{2-10}
                                            & \multicolumn{3}{c|}{\textbf{NRD}}                                                  & \multicolumn{3}{c|}{\textbf{BLT}}                                                  & \multicolumn{3}{c|}{\textbf{MLT}}                                                  \\ \cline{2-10} 
                                            & \multicolumn{1}{c|}{\textit{LT}} & \multicolumn{1}{c|}{\textit{G}} & \textit{LT+G} & \multicolumn{1}{c|}{\textit{LT}} & \multicolumn{1}{c|}{\textit{G}} & \textit{LT+G} & \multicolumn{1}{c|}{\textit{LT}} & \multicolumn{1}{c|}{\textit{G}} & \textit{LT+G} \\ \hline
\multicolumn{1}{|l|}{\textbf{GPT-4o}}       & \multicolumn{1}{c|}{0.87}        & \multicolumn{1}{c|}{0.89}       & 0.88          & \multicolumn{1}{c|}{0.88}        & \multicolumn{1}{c|}{0.89}       & 0.89          & \multicolumn{1}{c|}{0.84}        & \multicolumn{1}{c|}{0.88}       & 0.86          \\ \hline\hline

\multicolumn{1}{|l|}{\textbf{QWEN2.5 72b}}  & \multicolumn{1}{c|}{0.74}        & \multicolumn{1}{c|}{0.87}       & 0.81          & \multicolumn{1}{c|}{0.71}        & \multicolumn{1}{c|}{0.83}       & 0.77          & \multicolumn{1}{c|}{0.76}        & \multicolumn{1}{c|}{0.87}       & 0.82          \\ \hline
\multicolumn{1}{|l|}{\textbf{LLaMa3.1 70b}} & \multicolumn{1}{c|}{0.72}        & \multicolumn{1}{c|}{0.82}       & 0.77          & \multicolumn{1}{c|}{0.72}        & \multicolumn{1}{c|}{0.81}       & 0.76          & \multicolumn{1}{c|}{0.72}        & \multicolumn{1}{c|}{0.81}       & 0.77          \\ \hline\hline

\multicolumn{1}{|l|}{\textbf{M Nemo 12b}}   & \multicolumn{1}{c|}{0.36}        & \multicolumn{1}{c|}{0.49}       & 0.43          & \multicolumn{1}{c|}{0.36}        & \multicolumn{1}{c|}{0.42}       & 0.39          & \multicolumn{1}{c|}{0.41}        & \multicolumn{1}{c|}{0.55}       & 0.48          \\ \hline\hline

\multicolumn{1}{|l|}{\textbf{LLaMa3.1 8b}}  & \multicolumn{1}{c|}{0.47}        & \multicolumn{1}{c|}{0.62}       & 0.54          & \multicolumn{1}{c|}{0.44}        & \multicolumn{1}{c|}{0.57}       & 0.50          & \multicolumn{1}{c|}{0.50}        & \multicolumn{1}{c|}{0.66}       & 0.58          \\ \hline
\multicolumn{1}{|l|}{\textbf{LLaMa3 8b}}    & \multicolumn{1}{c|}{0.45}        & \multicolumn{1}{c|}{0.48}       & 0.46          & \multicolumn{1}{c|}{0.39}        & \multicolumn{1}{c|}{0.40}       & 0.40          & \multicolumn{1}{c|}{0.48}        & \multicolumn{1}{c|}{0.53}       & 0.50          \\ \hline
\multicolumn{1}{|l|}{\textbf{QWEN2.5 7b}}   & \multicolumn{1}{c|}{0.49}        & \multicolumn{1}{c|}{0.62}       & 0.56          & \multicolumn{1}{c|}{0.46}        & \multicolumn{1}{c|}{0.48}       & 0.47          & \multicolumn{1}{c|}{0.58}        & \multicolumn{1}{c|}{0.73}       & 0.65          \\ \hline\hline

\multicolumn{1}{|l|}{\textbf{LLaMa3.2 3b}}  & \multicolumn{1}{c|}{0.40}        & \multicolumn{1}{c|}{0.50}       & 0.45          & \multicolumn{1}{c|}{0.34}        & \multicolumn{1}{c|}{0.33}       & 0.34          & \multicolumn{1}{c|}{0.42}        & \multicolumn{1}{c|}{0.47}       & 0.45          \\ \hline
\end{tabular}
\caption{Average accuracy results per language group and model. \textbf{NRD} stands for Nordic, \textbf{BLT} stands for Baltic, and \textbf{MLT} stands for multilingual language groups. \textit{LT-R}, \textit{G}, and \textit{LT+G} stand for Lithuania-related history questions, general history questions and merged history questions respectively. \textbf{M Nemo 12b} refers to Mistral Nemo 12b model.}
\label{tab:avg_results}
\end{table*}

Our results demonstrated that all the models except for GPT-4o obtained better scores for general history questions rather than for LT-related ones. This observation is expected due to a biased training datasets for such models towards English-centric data. 

The largest evaluated model, GPT-4o, performed consistently better than other models for LT-related and general history questions in all language groups. The model achieved a maximum average score of 0.88 for LT-related history questions for the Baltic group (BLT) and performed similarly for Nordic languages (NRD) with a score of 0.87, though it showed slightly weaker performance in the multilingual group (MLT), scoring 0.84. These results suggest better knowledge representation for Nordic and Baltic language groups in LT-related history exams. Among the individual languages, English and Lithuanian were the best-performing languages for both LT-related and general history questions.

The 70b group of models (QWEN2.5 72b and LLaMa3.1 70b) demonstrated second best performance across all the types of questions. QWEN2.5 showed lower accuracy for Baltic languages on average, obtaining similar scores for MLT and NRD groups. Also, in both types of questions, QWEN2.5 showed similar trends of receiving lower scores in Estonian and Latvian, but higher scores for Nordic languages. Additionally, its performance was better in general questions across all languages, but when it comes to the alignment with LT-related, model was able to output better results for English, Swedish, and Danish rather than for Lithuanian or Baltic languages. In contrast, LLaMa3.1 70b did not performed at par with diff language groups. The results are similar for all languages in all questions with Arabic being the weakest and Lithuanian with English slightly stronger than others.

\begin{table*}[]
\centering
\begin{tabular}{|l|ccc|ccc|ccc|}
\hline
                                  & \multicolumn{3}{c|}{\textbf{SW}}                                                     & \multicolumn{3}{c|}{\textbf{DN}}                                                     & \multicolumn{3}{c|}{\textbf{EN}}                                                     \\ \hline
                                  & \multicolumn{1}{c|}{\textit{LTR}} & \multicolumn{1}{c|}{\textit{G}} & \textit{LTR+G} & \multicolumn{1}{c|}{\textit{LTR}} & \multicolumn{1}{c|}{\textit{G}} & \textit{LTR+G} & \multicolumn{1}{c|}{\textit{LTR}} & \multicolumn{1}{c|}{\textit{G}} & \textit{LTR+G} \\ \hline
\textbf{NRD LLaMa3 8b}    & \multicolumn{1}{c|}{0.43}          & \multicolumn{1}{c|}{0.50}       & 0.46          & \multicolumn{1}{c|}{0.42}          & \multicolumn{1}{c|}{0.51}       & 0.46          & \multicolumn{1}{c|}{$-$}          & \multicolumn{1}{c|}{$-$}       & $-$          \\ \hline
\textbf{GPT-SW3 126m}    & \multicolumn{1}{c|}{0.27}          & \multicolumn{1}{c|}{0.22}       & 0.24          & \multicolumn{1}{c|}{0.27}          & \multicolumn{1}{c|}{0.21}       & 0.24          & \multicolumn{1}{c|}{0.27}          & \multicolumn{1}{c|}{0.20}       & 0.24          \\ \hline
\textbf{GPT-SW3 356m}    & \multicolumn{1}{c|}{0.27}          & \multicolumn{1}{c|}{0.21}       & 0.24          & \multicolumn{1}{c|}{0.27}          & \multicolumn{1}{c|}{0.21}       & 0.24          & \multicolumn{1}{c|}{0.23}          & \multicolumn{1}{c|}{0.20}       & 0.21          \\ \hline
\textbf{GPT-SW3 1.3b}    & \multicolumn{1}{c|}{0.23}          & \multicolumn{1}{c|}{0.24}       & 0.23          & \multicolumn{1}{c|}{0.23}          & \multicolumn{1}{c|}{0.23}       & 0.23          & \multicolumn{1}{c|}{0.23}          & \multicolumn{1}{c|}{0.24}       & 0.24          \\ \hline
\textbf{GPT-SW3 6.7b} & \multicolumn{1}{c|}{0.32}          & \multicolumn{1}{c|}{0.27}       & 0.29          & \multicolumn{1}{c|}{0.33}          & \multicolumn{1}{c|}{0.29}       & 0.29          & \multicolumn{1}{c|}{0.24}          & \multicolumn{1}{c|}{0.28}       & 0.26          \\ \hline


\end{tabular}

\caption{Accuracy results for Nordic fine-tuned models. \textbf{NRD LLaMa3 8b} refers to pre-trained LLaMa3 8b by AI Sweden. \textit{LTR}, \textit{G}, and \textit{LTR+G} stand for Lithuania-related history questions, general history questions and merged history questions respectively. \textbf{SW} (Swedish), \textbf{DN} (Danish), \textbf{EN} (English) indicate a language that was used for evaluating the model.}
\label{tab:nordic_models_res}
\end{table*}
In case of Mistral Nemo 12b, the model scored the smallest scores comparing to other, even smaller (7-8b, 3b) models. It showed similar results across all language groups, obtaining the same average accuracy scores (36\%) on Baltic and Nordic group on LT-related questions and a better performance for Nordic group on general questions than for Baltic. The average of MLT group was better, even though neither score was higher than 64\%. 

LLaMa3 8b, QWEN2.5 7b, and LLaMa3.1 8b demonstrated a weaker performance when tested on BLT group across all questions. Using Lithuanian showed a better results. Similarly, Swedish and Danish helped QWEN2.5 7b obtain a better score. This results indicate that these models are better aligned with Lithuanian national history when asked in a language from Nordic group or in Lithuanian. LLaMa3.2 3b showed similar performance on NRD group to Mistral Nemo, but in MLT and BLT settings it received the lowest scores. 

The Nordic-specific models performed similarly on all their supported languages. From the considered models, NRD LLaMa3 is a clear winner. It demonstrated a similar performance across its supported languages and is very close to LLaMa3.2 performance on Swedish and Danish, but still underperformed LLaMa3.1 8b and QWEN2.5 7b on the corresponding languages. When it comes to a family of GPT-SW3, the greater the amount of parameters - the better performance. GPT-SW3 6.7b outperformed other versions of the model across Swedish and Danish. However, on English, GPT-SW3 with 126m performed better on LT-related questions.

In conclusion, our experiments show that GPT-4o performs consistently better across all tested languages and language groups on LT-related and general history questions. Larger open source models, LLaMa3.1 70b and QWEN2.5 72b also performed consistently well in all languages. Mistral Nemo 12b, LLaMa3 8b, LLaMa3.2 3b, and LLaMa3.1 8b demonstrated significant gaps in their historical knowledge for LT-related history questions within Baltic language group, even when Lithuanian is part of this group. The better performance was obtained in Nordic language group, indicating that cultural or historical alignment alone does not guarantee higher accuracy for these models. The Nordic pre-trained models were not able to outperform the multilingual model, rejecting our initial hypothesis.

\section{Conclusion}

This study evaluated the performance of Large Language Models (LLMs) on Lithuanian historical multiple-choice question-answering tasks, focusing on Baltic, Nordic, and other language groups. The models were evaluated on the Lithuanian national history related (LT-related) questions and a general history questions.

Our findings showed that GPT-4o consistently outperformed all other tested models across languages, achieving the highest scores for LT-related and general history questions, with slightly better results for Baltic and Nordic languages. Among open-source models, larger models QWEN2.5 72b and LLaMa3.1 70b performed well but did not match GPT-4o, especially in Baltic languages. Smaller models, including Mistral Nemo 12b, LLaMa3.2 3b, QWEN 7B, LLaMa3.1 8B, and LLaMa3 8b demonstrated weaker results with Baltic languages, including Lithuanian, while performing better in Nordic and multilingual groups.  

Nordic fine-tuned models performed consistently across their supported languages but failed to surpass general multilingual models, even within their specialized domain. These findings highlight that shared cultural or historical context alone does not guarantee better model performance. To bridge these gaps, further efforts are needed to develop targeted datasets and fine-tuning strategies to improve LLM alignment with less-resourced languages like those in the Baltic language group.

\bibliographystyle{acl_natbib}
\bibliography{nodalida2025}

\appendix
\section{Manual Translation Quality Evaluation} \label{sec:lt-en-tr}
\begin{table*}
    \centering
    
    \begin{tabular}{|c|c|c|c|c|c|c|}
    \hline
        Lang Pair & \# Reject (A) & \# Reject (B) &  \# Accept (A) & \# Accept (B) & Intersect, \%\\ \hline
        LT-EN & 11 & 18 &  89 & 82  & 75 \\ \hline
        EN-UA & 10 & 30 &  90 & 70 & 75 \\ \hline
        EN-AR & 28 & 21 &  72 & 79 & 65 \\ \hline
        LT-EST* & 13 & 54 &  87 & 46 &  0.55 \\ \hline
        EN-SW & 1 & 6 & 99 & 94  & 0.9 \\ \hline
        EN-DN & 9 & 41 & 91 & 59 & 0.6  \\ \hline
        EN-FN & 15 & 33 & 78  & 67 & 0.73 \\ \hline
        LT-LAV* & 27 & 43  &  73 & 57 & 0.7 \\ \hline
    \end{tabular}
    \caption{Annotation results. * indicates translation with DeepL from Lithuanian to the target language. \# Reject and Accept refer to a number of rejected and accepted samples by the annotator (marked with letters A and B). Intersect indicates a percentage of samples that annotators assigned the same label.}
    \label{tab:agrr}
\end{table*}

The annotation guidelines and examples can be found in our GitHub repository~\footnote{HIDDEN FOR REVIEW}. For each translated language, we utilized the same following strategy. We recruited native speaker annotators, who are also are proficient in English. They were presented with 80 random samples from the dataset distinct for each annotator and 20 samples that are the same for each annotator. From those 80 samples, 40 were selected from a pool of Lithuanian history questions, and other 40 from the general history question. The same approach was applied for the remaining 20 samples: 10 were selected from a pool of Lithuanian history questions, and other 10 from the general history question.

The annotators were presented with the translated English question, its answer choices and the corresponding translation for the question and choices. in the case of Lithuanian to English translation, the pairs of Lithuanian and English were presented. They were instructed to determine if the translation is correct from the following standpoints. The translation accurately conveys the meaning of the English or Lithuanian text. The order of answers (with respect to the letters) is the same in both languages. The names of historical figures, locations, dates, or events are correctly translated and align with conventions.  Text semantics are clear and do not change the intent or emphasis of the question or answers. If the translation contains grammar or phrasing issues, or minor typos, they do not lead to confusion or ambiguity and do not change the semantics.

If the translation does not fit the requirements above, the translation is rejected. The annotation results and agreements (in a form of number of intersections) are presented in the Table~\ref{tab:agrr}. During our experiments, chatGPT showed poor results when translating to Latvian and Estonian. Therefore, we used DeepL to translated Lithuanian to Latvian and Estonian. The annotation in the Table~\ref{tab:agrr} corresponds to DeepL translation.

\end{document}